%% file: main.tex
\documentclass[10pt,twocolumn,letterpaper]{article}

\usepackage{iccv}      % To produce the REVIEW version
\usepackage{graphicx}
\usepackage{algorithm}
\usepackage{algorithmic}
\usepackage[algo2e,linesnumbered,noend]{algorithm2e}
\usepackage{amsmath,bm}
\usepackage{amssymb}
\usepackage{multirow}
\usepackage{booktabs}
\usepackage{color}
\usepackage{float}
\usepackage{arydshln}
\usepackage{tabularx}
\usepackage{comment}
\usepackage{fontawesome}
\usepackage[T1]{fontenc}
\usepackage[accsupp]{axessibility} 
\DeclareMathOperator*{\argmin}{\arg\min} 

\usepackage{mathtools}
\usepackage{color, colortbl}
\definecolor{orange}{rgb}{0.988, 0.878, 0.533}

\usepackage{enumitem}

\definecolor{iccvblue}{rgb}{0.21,0.49,0.74}
\usepackage[pagebackref,breaklinks,colorlinks,allcolors=iccvblue]{hyperref}

\DeclarePairedDelimiterX{\infdivx}[2]{(}{)}{%
  #1\;\delimsize\|\;#2%
}

\usepackage[capitalize]{cleveref}
\crefname{section}{Sec.}{Secs.}
\Crefname{section}{Section}{Sections}
\Crefname{table}{Table}{Tables}
\crefname{table}{Tab.}{Tabs.}

\def\paperID{8160} % *** Enter the Paper ID here
\def\confName{ICCV}
\def\confYear{2025}

\makeatletter
\DeclareRobustCommand\onedot{\futurelet\@let@token\@onedot}
\def\@onedot{\ifx\@let@token.\else.\null\fi\xspace}
\def\eg{\emph{e.g}\onedot} 
\def\ie{\emph{i.e}\onedot}

\makeatother

\DeclareRobustCommand*{\IEEEauthorrefmark}[1]{%
    \raisebox{0pt}[0pt][0pt]{\textsuperscript{\footnotesize\ensuremath{#1}}}}

\newcommand{\mathbbm}[1]{\text{\usefont{U}{bbm}{m}{n}#1}}

\newcommand{\green}[1]{\textcolor[rgb]{0,0.5,0}{#1}}
\newcommand{\blue}[1]{\textcolor[rgb]{0.21,0.49,0.74}{#1}}

\begin{document}

\title{Unlearning the Noisy Correspondence Makes CLIP More Robust}
\author{\centering
Haochen Han\IEEEauthorrefmark{1}\ ,
Alex Jinpeng Wang\IEEEauthorrefmark1\IEEEauthorrefmark2$^{\dagger}$,
Peijun Ye\IEEEauthorrefmark1, 
Fangming Liu\IEEEauthorrefmark{1}$^{\dagger}$\
\and
{ \IEEEauthorrefmark1 Peng Cheng Laboratory}
{ \IEEEauthorrefmark2 Central South University}
\and
\texttt{\normalsize \faEnvelopeO:hhc2077@outlook.com; \faGithubSquare:\href{https://github.com/hhc1997/NCU}{https://github.com/hhc1997/NCU}}}

% \and
\maketitle
% \begin{comment}
%%%%%%%%% ABSTRACT
\begin{abstract}

The data appetite for Vision-Language Models (VLMs) has continuously scaled up from the early millions to billions today, which faces an untenable trade-off with data quality and inevitably introduces Noisy Correspondence (NC) samples. Undoubtedly, such semantically unrelated data significantly impairs the performance of VLMs. Previous efforts mainly address this challenge by estimating refined alignment for more precise guidance. However, such resource-intensive pipelines that train VLMs from scratch struggle to meet realistic data demands. In this paper, we present a brand new perspective that seeks to directly eliminate the harmful effects of NC in pre-trained VLMs. Specifically, we propose NCU, a Noisy Correspondence Unlearning fine-tuning framework that efficiently enhances VLMs' robustness by forgetting learned noisy knowledge. The key to NCU is learning the hardest negative information, which can provide explicit unlearning direction for both false positives and false negatives. Such twin goals unlearning process can be formalized into one unified optimal transport objective for fast fine-tuning. We validate our approach with the prevailing CLIP model over various downstream tasks. Remarkably, NCU surpasses the robust pre-trained method on zero-shot transfer while with lower computational overhead. The code will be released upon acceptance.

\let\thefootnote\relax\footnotemark\footnotetext{$^{\dagger}$ Co-corresponding authors.}

\end{abstract}

\section{Introduction}
The pursuit of general intelligence has driven progress in multimodal learning, which seeks to integrate and understand multiple sensory modalities like humans. Large-scale vision-language training, exemplified by CLIP \cite{radford2021learning}, is seen as a key milestone in multimodal learning due to its remarkable transfer capabilities in real-world applications, such as image-text retrieval \cite{jia2021scaling, goel2022cyclip, pateltripletclip} and robotics control \cite{shridhar2022cliport}.

However, much of their success can be attributed to scaling laws enabled by massive training data. As every coin has two sides, the insatiable demand for data forces a difficult trade-off between quantity and quality, which inevitably introduces noisy correspondence into the training set. Taking the CC3M dataset \cite{sharma2018conceptual} as an example, despite being filtered from 500 million images, it still contains at least 3\% \cite{huang2021learning} unrelated image-text pairs, \ie, \textit{false positive}. To make matters worse, training on massive data necessitates larger batch sizes (32K used in CLIP), which increases the likelihood of unpaired samples sharing semantic similarities, \ie, \textit{false negative}. Undoubtedly, such two-aspects noisy correspondence can significantly impair the performance of vision-language models.

\begin{figure}[t]
\centering  
\includegraphics[width=\columnwidth]{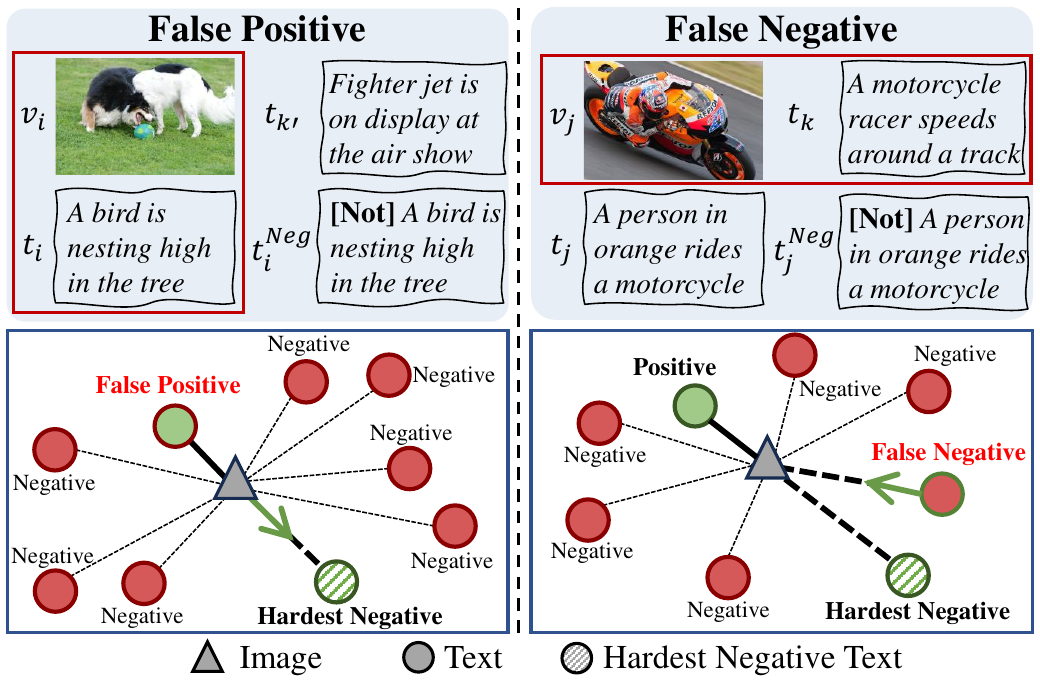}
\caption{\textbf{Illustration on the core concept of NCU.} The twin goals unlearning process is guided by the learned hardest negative information. For the FP, $t_i^{\scriptstyle Neg}$ directly pulls $v_i$ away from the mismatched $t_i$. While for the FN, $t_i^{\scriptstyle Neg}$ acts as a distance upper bound to facilitate modeling many-to-many relations.}
\label{fig:intro}
% \vspace{-0.2cm} 
\end{figure}

To endow robustness against NC, one natural direction is to revise the pre-training paradigm \cite{gao2022pyramidclip, andonian2022robust, gao2024softclip, bulat2024fff, huang2024noise} that supervises VLMs with refined alignment. However, existing methods require training from scratch and may rely on guidance from external large models \cite{gao2024softclip, bulat2024fff}. Such resource-intensive pipelines obviously struggle to face the realistic demand, especially with today's billion-scale datasets \cite{schuhmann2021laion}. Hence, it is necessary to address the NC problem in vision-language training with a cost-effective method.

In this paper, we think outside the box of robust pre-training and pose an important question: \textit{Can we directly eliminate the harmful effects of NC in pre-trained VLMs?} To answer this question, we resort to machine unlearning \cite{bourtoule2021machine} and present NCU, a \textbf{N}oisy \textbf{C}orrespondence \textbf{U}nlearning fine-tuning framework that improves the robustness of CLIP by erasing learned noisy knowledge. Machine unlearning is a reversed learning process that aims to delete the influence of specific training samples from trained models. Despite its promise in widespread tasks \cite{foster2024fast,fan2024salun}, unlearning the NC in VLMs remains unexplored due to a key challenge: the ambiguous forgetting direction would corrupt the learned semantic structure in the feature space. To address this, we propose to learn the hardest negative information that can provide explicit unlearning direction. As illustrated in \cref{fig:intro}, on the one hand, the negative information would directly serve as reliable supervision for forgetting false positives. On the other hand, it would facilitate the modeling of many-to-many relationships among unpaired data for forgetting false negatives. Then, we show that such twin goals unlearning process can be formalized as one unified optimal transport problem, which efficiently fine-tunes CLIP to resist both FP and FN.

Our main contributions are highlighted below:
\begin{itemize}
    \item To the best of our knowledge, this work could be the first study to eliminate the harmful effects of noisy correspondence from pre-trained CLIP. 
    \item We propose the NCU framework, which efficiently unlearns FP and FN with explicit direction derived from the hardest negative information.
    \item We demonstrate that NCU achieves significant improvements over CLIP on several downstream tasks and surpasses the previous robust pre-training method with lower computational overhead.
\end{itemize}

\section{Related Work}
\label{sec:related work} 

\paragraph{Noisy Correspondence Learning.} Noisy correspondence refers to the alignment error presented in multimodal data. The false positive is a typical NC problem, where irrelevant multimodal pairs are wrongly treated as matched. To alleviate this, several techniques have been developed in various multimodal applications, including cross-modal retrieval \cite{huang2021learning, hu2023cross, qin2024cross, han2024learning}, video temporal learning \cite{han2022temporal, lin2023multi}, multimodal person re-identification \cite{yang2024robust,qin2024noisy}, question answering \cite{jiang2023robust}, and image captioning \cite{kang2023noise, fu2024noise}. In more complex scenarios, \eg, vision-language pre-training \cite{huang2023nlip, huang2024noise}, models also suffer from false negatives caused by the training paradigm \cite{gao2022pyramidclip}, where similar unpaired samples are forced to be distant. Considering the computational burden of large VLMs, this work presents a low-carbon solution to directly improve the robustness of pre-trained VLMs.

\paragraph{Contrastive Vision Language Models.} Contrastive vision language models (VLMs) \cite{mu2022slip, goel2022cyclip, gu2024rwkv, wang2022eclip, yaofilip, fan2024improving} aim to learn visual representations by the corresponding textual supervision, which have attracted significant attention due to their simplicity and powerful representation capability. Pioneering works CLIP \cite{radford2021learning} and ALIGN \cite{jia2021scaling} have shown great success via learning from massive image-text pairs. However, such web-crawled data are noisy \cite{thomee2016yfcc100m, schuhmann2021laion} and inevitably harm the efficacy of existing VLMs. To tackle this issue, a series of works attempted to train the VLM with refined soft image-text alignments by label smoothing \cite{gao2022pyramidclip}, knowledge distillation \cite{andonian2022robust}, fine-grained intra-modal guidance \cite{gao2024softclip}, text rewriting \cite{bulat2024fff}, or positive-negative contrastive loss \cite{huang2024noise}. Besides, OT-based methods\cite{wudata, shiot} have also emerged as they naturally model such matching problems. Despite the success, previous works focus on training robust VLMs from scratch, which overlooks readily available pre-trained models and incurs unnecessary computational costs. To this end, this paper pioneers an efficient approach to enhance model robustness by unlearning noisy information from pre-trained models.

\paragraph{Machine Unlearning.} Recent advances in MU mainly focus on practical approximate unlearning, which seeks to mimic the behavior of a model re-trained from scratch. Driven by privacy concerns, existing MU works \cite{mehta2022deep, foster2024fast, fan2024salun} in computer vision focus on image classification that attempts to forget specific classes. In parallel, MU has also become a popular topic in large language models due to its capability to eliminate harmful responses \cite{zhang2024negative,liu2024large}. However, multimodal forgetting remains under-explored in the literature. Pioneering works \cite{Kravets_2025_WACV, kravets2025zeroshot} studied data-free class removal for CLIP's downstream image classification. To date, none of the existing MU methods has explored the unlearning of noisy concepts from VLMs.

%%Let's start
\section{Preliminaries}
\subsection{Contrastive Language-Image Pre-training}
CLIP is a vision-language model trained on millions of web-harvested image-text pairs. We consider a batch of $N$ image-text pairs $\{ v_i, t_i\}_{i=1}^N$ sampled from a cross-modal dataset $\mathbbm{D}$, where $v_i$ and $t_i$ represent the raw image and corresponding text, respectively. The goal of CLIP is to train two modality-specific encoders that bring matched pairs closer while pushing unmatched ones apart. Specifically, image embedding $\bm{v}_i \in \mathbb{R}^d$ and text embedding $\bm{t}_i \in \mathbb{R}^d$ are obtained by passing $v_i$ and $t_i$ through the image encoder $f_v$ and text encoder $f_t$, respectively, where $d$ is the embedding dimension. The encoded $l_2$ normalized embeddings are then aligned in the feature space by minimizing the contrastive objective, \ie, InfoNCE loss:
\begin{equation}\label{eq: InfoNCE}
\mathcal{L}_{v \to t}^{CL} = -\frac{1}{N} \sum_{i=1}^N \log \frac{\exp{(\langle \bm{v}_i, \bm{t}_i \rangle / \tau)}}{\sum_{j=1}^N \exp{(\langle \bm{v}_i, \bm{t}_j \rangle / \tau)}},
\end{equation}
where $\langle \cdot \rangle$ represents the inner product and $\tau$ is a trainable temperature parameter. As InfoNCE loss is symmetric, we can define $\mathcal{L}_{t \to v}^{CL}$ similarly. The complete CLIP training objective is formulated as: $\mathcal{L}_{CLIP} = \mathcal{L}_{v \to t}^{CL} + \mathcal{L}_{t \to v}^{CL}$.

Despite its promising performance, the standard contrastive learning can suffer from the noisy correspondence problem in two aspects. First, the web-collected pairs inevitably contain an unknown portion of mismatched data, \ie, false positives. Second, hard target alignment neglects the potential semantic similarity among unpaired samples, \ie, false negatives, especially under large batch settings.

\subsection{Machine Unlearning}
Given a CLIP model (also named \emph{reference model}) $\{f_v, f_t\}$ that is already trained on a cross-modal dataset $\mathbb{D}$, machine unlearning aims to fine-tune the model to forget a specific subset $\mathbbm{D}_{FG} \subseteq \mathbbm{D}$ while maintaining effectiveness on the retained set $\mathbbm{D}_{RT} =  \mathbbm{D} \setminus \mathbbm{D}_{FG}$. Ideally, the model should behave as if it were trained without any sample from $\mathbbm{D}_{FG}$. In principle, re-training the model from scratch on $\mathbbm{D}_{RT}$ would serve as the gold standard. However, since CLIP is trained on massive-scale data, it is unrealistic to obtain a forget set that includes all noisy information, especially when some data are not publicly accessible. Therefore, we focus on an approximate unlearning approach in which $\mathbbm{D} = \mathbbm{D}_{FG}\,\cup\,\mathbbm{D}_{RT}$ does not need to contain all training pairs, making the unlearning process more practical for real-world scenarios.

The most straightforward method to unlearn is \textit{gradient ascent} or its variants, which optimizes the negative prediction loss over the forget set. Another typical approach is performing \textit{forget loss} that encourages the model to relearn the modified form of undesired data. For example, we can update CLIP by minimizing InfoNCE loss in pair $\{ v_i, \tilde{t}_i\} \sim \mathbbm{D}_{FG}$ to forget the relation between $v_i$ and $t_i$, where $\tilde{t}_i \neq t_i$ could be random or hand-crafted text to replace the original. Based on these, existing MU methods have shown promising progress in class forgetting and LLM privacy protection. However, applying these strategies to CLIP unlearning poses a key challenge: the ambiguous forgetting direction would corrupt the learned semantic structure in the feature space. In other words, the model forgets the undesired data by learning other meaningless patterns.

\section{Methodology}
To tackle the above issues, we introduce the Noisy Correspondence Unlearning (NCU) framework. In the following, we first introduce the division of forget and retained sets in Sec~\ref{subsec: data_splite}. Subsequently, we elaborate on learning the hardest negative information in Sec~\ref{subsec: learn_hn} and explain how to formalize the twin goals unlearning process into an optimal transport object for efficiently fine-tuning in Sec~\ref{sec:related work}. The overall training pseudo-code is shown in Supplementary \blue{A}.

\subsection{Identifying the Forget Set}\label{subsec: data_splite}
Unlike standard MU tasks with a predefined forget set, we need to manually identify mismatched samples from CLIP's training data to construct it. As pre-trained CLIP has shown strong representation capability, we propose using the basic similarity score to obtain $\mathbbm{D}_{FG}$ and $\mathbbm{D}_{RT}$, \ie,
\begin{equation}\label{eq: sim_score}
\small
\omega_i = \frac{1}{2}\left[\frac{\exp{(\big<\bm{v}_i, \bm{t}_i \big>/\tau)}}{\sum_{j=1}^N\exp{(\big<\bm{v}_i, \bm{t}_j\big>/\tau)}} + \frac{\exp{(\big<\bm{t}_i, \bm{v}_i \big>/\tau)}}{\sum_{j=1}^N\exp{(\big<\bm{t}_i, \bm{v}_j\big>/\tau})}\right].
\end{equation}
By comparing $(v_i, t_i)$ with other cross-modal samples in the batch, $\omega_i$ serves as a clean confidence that measures the extent of semantic match. Then, we select pairs with the lowest $P\%$ of $\omega_i$ within the batch as false positives to construct the forget set $\mathbbm{D}_{FG}$, while treating the remaining in-batch pairs as the retained set $\mathbbm{D}_{RT}$.

Note that $\mathbbm{D}_{FG}$ and $\mathbbm{D}_{RT}$ are dynamically selected at each batch, which enjoys two merits: 1) CLIP could be efficiently updated with one intra-batch optimization; 2) The forget-retain ratio could be flexibly adjusted through the predefined parameter $P$. 

\begin{figure*}[t]
\centering  
\includegraphics[width=0.9\textwidth]{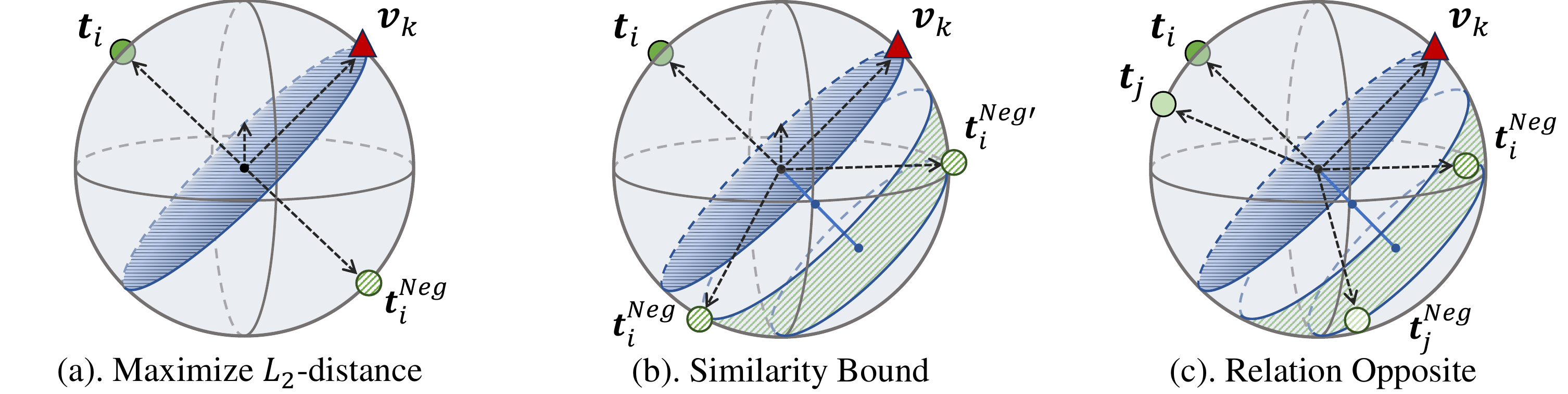}
\caption{\textbf{Illustration of the optimization objective to learn the hardest negative semantics.} (a) Previous attempts that directly maximize the $L_2$ distance prevent $\bm{t}_i^{\scriptstyle Neg}$ from providing certain guidance for unpaired images, \eg, $\bm{v}_k$. (b) We bound the similarity with margins for a more relaxed semantic separation, but it may lead to uncertain targets, \eg, $\bm{t}_i^{\scriptstyle Neg}$ and $\bm{t}_i^{\scriptstyle Neg\prime}$. (c) We further preserve relation structures for precise objectives. The intuition is that the opposite text should also maintain semantic relationships, \eg, $\langle \bm{t}_i, \bm{t}_j^{\scriptstyle Neg} \rangle \approx \langle \bm{t}_j, \bm{t}_i^{\scriptstyle Neg} \rangle$.}
\label{fig: illustration}
% \vspace{-0.1cm} 
\end{figure*}

\subsection{Learning Hardest Negative Semantics}\label{subsec: learn_hn}
To guide CLIP with an explicit unlearning direction, we aim to learn the hardest negative semantics as supervision. Intuitionally, for an irrelevant pair $(v_i, t_i)$ that misleads the model with \textit{`$v_i$ and $t_i$ are matched'}, we encourage the model to forget this information by relearning that \textit{`$v_i$ and $t_i$ are not matched'}. From a data utilization viewpoint, this paradigm is similar to negative learning\cite{kim2021joint} that supervises the model with complementary information \cite{hu2023cross}, \ie, pushing the candidate away from other unpaired samples. Differently, our method seeks the hardest negative information to avoid uncertain optimization directions. 

To achieve this, we incorporate a set of learnable vectors to represent the textual negative semantics inspired by prompt learning \cite{zhou2022learning}. Specifically, for any training pair $(v_i, t_i)$, the token features of $t_i$ are combined with $m$ shared prompt vectors to present the corresponding negative semantics $t_i^{\scriptstyle  Neg}$ in the feature space. While such prompt-driven semantic negation of CLIP has demonstrated success in out-of-distribution detection \cite{wang2023clipn, li2024learning}, existing methods are confined to closed-set downstream tasks with limited category concepts. In contrast, our challenge lies in extending the semantic opposite operation into the open-set knowledge that CLIP pre-trained.

Intuitively, the hardest negative satisfies two constraints in the feature space: 1) $t_i^{\scriptstyle Neg}$ needs to maximize its distance from $v_i$ and $t_i$; 2) $t_i^{\scriptstyle Neg}$ should maintain certain similarity to those unpaired images, as it is not a wrong description \cite{wang2023clipn} despite being semantically irrelevant to $v_j$. Furthermore, we only use $\mathbbm{D}_{RT}$ to learn the prompt tokens to avoid overfitting caused by noisy correspondence. For notation simplicity, we denote $\tilde N$ as the size of $\mathbbm{D}_{RT}$ within each batch. Based on the above insights, we propose the following intra-modal and cross-modal training objectives.

\paragraph{Text Relation Opposite.}It encourages semantic separation between the embeddings of negative text and its original. Most previous works \cite{wang2023clipn, li2024learning} typically reduce the per-instance similarity gap among textual pairs, \eg, $\|\bm{t}_i - \bm{t}_i^{\scriptstyle Neg}\|_2 \rightarrow 2$ \cite{wang2023clipn} to directly maximize its $L_2$ distance. However, such rigid constraint incurs a crucial limitation in the open-set semantic space---enforcing maximal distance pushes $t_i^{\scriptstyle Neg}$ away from unpaired images (\cref{fig: illustration}(a)), which is contrary to our objective. To this end, we propose a relaxed similarity bound to constrain the semantic separation:
\begin{equation}\label{eq: sep_loss}
\mathcal{L}^{sep} = \frac{1}{\tilde N} \sum_{i=1}^{\tilde N} \big([\alpha - \langle \bm{t}_i, \bm{t}_i^{\scriptstyle Neg} \rangle]_{+} +[\langle \bm{t}_i, \bm{t}_i^{\scriptstyle Neg}\rangle - \beta ]_{+}\big),
\end{equation}
where $\alpha < 0$ and $\beta < 0$ are the margin parameters  to locate $\langle \bm{t}_i, \bm{t}_i^{\scriptstyle Neg}\rangle \in [\alpha, \beta]$, and $[x]_+ = max(x,0)$ is the hinge function. As illustrated in \cref{fig: illustration}(b), although minimizing Eq.\eqref{eq: sep_loss} enables ${t}_i^{\scriptstyle Neg}$ to be distant from $t_i$ while remaining relatively similar to unpaired images, the broad range of variations makes training convergence difficult. To address this issue, we propose to perform semantic opposite at the relation level instead of the instance level, which is achieved by preserving the geometrical structures among all negative and original text within the batch:
\begin{equation}\label{eq: rel_loss}
\mathcal{L}^{rel} = \frac{1}{\tilde N} \sum_{i=1}^{\tilde N} \sum_{j=1}^{\tilde N} \big( \langle \bm{t}_i, \bm{t}_j^{\scriptstyle Neg} \rangle - \langle \bm{t}_j, \bm{t}_i^{\scriptstyle Neg} \rangle \big)^2.
\end{equation}
As shown in \cref{fig: illustration}(c), regularizing the negative-original relation consistency can guide $t_i^{\scriptstyle Neg}$ toward a precise location in the feature space.

\paragraph{Image-text Matching Opposite.}It aims to model the alignment between the embeddings of negative text and images. As discussed, $t_i^{\scriptstyle Neg}$ provides positive supervision to unpaired images while separating from its paired image, which presents opposite matching patterns to the normal contrastive objective. To achieve this, we take inspiration from Sigmoid loss \cite{zhai2023sigmoid} that efficiently supports such multi-positive alignment. Specifically, it guides per cross-modal pair independently by the binary matching target:
\begin{equation}\label{eq: im_loss}
\begin{aligned}
\mathcal{L}^{itm}  =  & \frac{1}{\tilde N} \sum_{i=1}^{\tilde N} \sum_{j=1}^{\tilde N}\big(\log \frac{1}{1 + \exp(m_{ij}(-\langle \bm{t}_i, \bm{v}_j \rangle/\tau))}
\\ + & \log \frac{1}{1 + \exp(-m_{ij}(-\langle \bm{t}_i^{\scriptstyle Neg}, \bm{v}_j \rangle/\tau))}
\big),
\end{aligned}
\end{equation}
where $m_{ij}$ equals $1$ for $i = j$ and $-1$ for $i \neq j$. In Eq.\eqref{eq: im_loss}, the first part follows the standard one-to-one matching to retain the original CLIP knowledge, while the second part utilizes the opposite binary target, \ie, $-m_{ij}$, to bring $t_i^{\scriptstyle Neg}$ closer to multiple images.

\begin{figure*}[t]
\centering  
\includegraphics[width=1\textwidth]{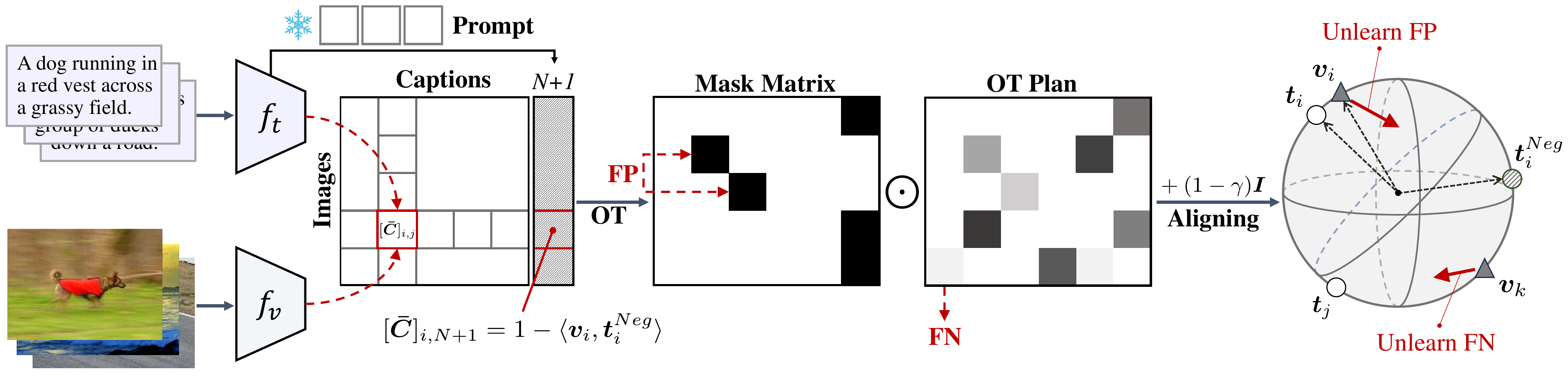}
\caption{\textbf{Overview of the Noisy Correspondence Unlearning process.} With the learned negative prompt frozen, we formulate an optimal transport problem guided by the hardest negative and then use the solved transport plan to robustly fine-tune the model $f_t$ and $f_v$.}
\label{fig: framework}
\end{figure*}

With the visual encoder frozen, the overall loss for learning the hardest negative semantics is balanced by a scaling factor $\lambda$ and given by:
\begin{equation}\label{eq: HN_loss}
\mathcal{L}^{HN} = \lambda (\mathcal{L}^{sep} + \mathcal{L}^{rel}) + \mathcal{L}^{itm}.
\end{equation}

\subsection{Hardest-Negative Guided Noise Unlearning}\label{subsec: unlearn}
The hardest negative semantics serve dual purposes in erasing the learned noisy correspondence: 1) guide CLIP to unlearn the false positive pair $(v_i, t_i)$ by matching $v_i$ with $t_i^{\scriptstyle Neg}$. 2) While for the well-matched pair $(v_i, t_i)$, $t_i^{\scriptstyle Neg}$ assists in inferring the soft alignment among unpaired data to unlearn the false negative pattern. To efficiently fine-tune CLIP, we formalize such twin goals unlearning process into one unified Optimal Transport problem.

\paragraph{Optimal Transport.} OT seeks to establish a flexible alignment between images and captions by computing a minimal-cost transport plan, where the cost refers to the expense of transporting mass from source to target distribution and is generally set to a distance measure \cite{gu2022keypoint}. Let $\bm{C} \in \mathbb{R}_{+}^{N \times N}$ denotes the cost matrix for the mini-batch, where $[\bm{C}]_{i,j} = 1 -\langle \bm{v}_i, \bm{t}_j \rangle$ is the cosine distance of $v_i$ and $t_j$. $\bm{\Gamma} \in \mathbb{R}_{+}^{N \times N}$ denotes the corresponding transport plan that $[\bm{\Gamma}]_{i,j}$ represents the alignment probability between $v_i$ and $t_j$. Formally, the objective of OT is defined as follows:
\begin{equation}\label{eq: ot}
\begin{aligned}
&\min_{\bm{\Gamma} \in \Pi(\bm{\mu},\bm{\nu})} \langle \bm{\Gamma} ,\bm{C} \rangle - \epsilon H(\bm{\Gamma})
\\\mbox{ s.t. } \Pi(\bm{\mu},\bm{\nu}) \! &= \! \{\bm{\Gamma} \! \in \! \mathbb{R}^{N\times N}_+ \vert \bm{\Gamma} \mathbbm{1}_{N} = \bm{\mu}, \bm{\Gamma}^{\top} \! \mathbbm{1}_{N} = \bm{\nu}\},
\end{aligned}
\end{equation}
where $\mathbbm{1}_{N}$ denotes a $N$-dimensional all-one vector, $\bm{\mu}, \bm{\nu} \in \mathbb{R}^N$ are probability measures representing the relative importance of each image and caption. Without prior knowledge, $\bm{\mu} = \frac{1}{N} \mathbbm{1}_{N}$ and $\bm{\nu} = \frac{1}{N} \mathbbm{1}_{N}$ are considered to be uniformly distributed since each pair is sampled independently. $H(\bm{\Gamma})$ is an additional entropy regularizer controlled by the smooth parameter $\epsilon$, which enables the OT objective to be solved by the rapid Sinkhorn-Knopp algorithm \cite{cuturi2013sinkhorn}.

\paragraph{Boosting OT via Hardest Negatives.}
To endow the transport plan with dual forgetting purposes, we reformulate Eq.\eqref{eq: ot} by imposing guidance from the hardest negative information. Specifically, for each image $v_i$, we extend its transport target from $\{t_i\}_{i=1}^N$ to include its paired negative text $t_i^{\scriptstyle Neg}$. As shown in \cref{fig: framework}, the negative text composes a new alignable column for the transport objective, which append the cost matrix $\bm{C}$ to $\bm{\bar{C}} \in \mathbb{R}_{+}^{N \times(N+1)}$, \ie,
\begin{equation*}\label{eq: cost_matrix}
[\bm{\bar{C}}]_{i,N+1} = 1 -\langle \bm{v}_i, \bm{t}_i^{\scriptstyle Neg} \rangle, [\bm{\bar{C}}]_{i,j} = [\bm{C}]_{i,j}, \forall i,j \in[1,N].
\end{equation*}
For the two parts $\mathbbm{D}_{FG}$ and $\mathbbm{D}_{RT}$ within each batch, the hardest negative should impose different guidance for distinct unlearning goals. To this end, we propose a mask-base constraint to the corresponding transport plan $\bm{\bar\Gamma}$ that regulates the effect of $t_i^{\scriptstyle Neg}$. Specifically, the mask matrix $\bm{M} \in \mathbb{R}_{+}^{N \times(N+1)}$ satisfies that
\begin{equation}\label{eq: mask_matrix}
[\bm{M}]_{i,j} = \begin{cases}
0, & \text{if } (v_i, t_i) \in \mathbbm{D}_{FG} \text{ and } j = i, \\
0, & \text{if } (v_i, t_i) \in \mathbbm{D}_{RT} \text{ and } j = N+1, \\
1, & \text{otherwise}.
\end{cases}
\end{equation}
For the toy example illustrated in \cref{fig: framework}, if the pair is considered to be mismatched, the transport mass between $v_i$ and $t_i$ should be constrained to zero. Conversely, for the well-matched pair, $t_i^{\scriptstyle Neg}$ acts as a lower limit where the transport mass between $v_i$ and $t_j$ should be higher than it. Following the solver from \cite{gu2022keypoint}, we model the mask constraint as the Hadamard product form that $\hat{\bm{\Gamma}} = \bm{M} \odot \bm{\bar\Gamma}$, and the optimal alignment is formulated as (detailed Sinkhorn solution is presented in Supplementary \blue{B}):
\begin{equation}\label{eq: masked_ot} 
\hat{\bm{\Gamma}}^* = \argmin_{\hat{\bm{\Gamma}} \in \Pi(\bm{\mu},\bm{\bar\nu})} \langle \hat{\bm{\Gamma}}  ,\bm{\bar{C}} \rangle - \epsilon H(\hat{\bm{\Gamma}} ),
\end{equation}
where $\bm{\bar\nu} = \frac{1}{N+1} \mathbbm{1}_{N+1}$ to satisfy the additional column.

\paragraph{The Unlearning Objective.}
Although $\hat{\bm{\Gamma}}^*$ provides the more refined alignment, we suggest further incorporating an identity-like matrix $\bm{I}$ for two merits. First,  diagonal elements are set as 1 for true positives to retain the initial alignment. Second, $[\bm{I}]_{i,N+1} = 1$ to enhance the unlearning for the possible false positive $(v_i, t_i) \in \mathbbm{D}_{FG}$. Thus, the overall alignment balanced by the factor $\gamma$ is defined as:
\begin{equation}\label{eq: alignment} 
 \bm{T} = \gamma \hat{\bm{\Gamma}}^* + (1 -\gamma)\bm{I}.
\end{equation}
To fine-tune CLIP with this soft alignment, we use the KL divergence to optimize the matching distribution. Formally, we denote the batched similarity matrix as $\bm{P} \in \mathbb{R}^{N \times(N+1)}$ where $\bm{P_i} = [\langle \bm{v}_i, \bm{t}_1 \rangle,\dots, \langle \bm{v}_i, \bm{t}_N \rangle, \langle \bm{v}_i, \bm{t}_i^{\scriptstyle Neg} \rangle ]^\top$. We obtain $\bm{ P_i}^{v2t}$ and $\bm{P_i}^{t2v}$ by applying row-wise and column-wise softmax operation to $\bm{ P}$, respectively. Correspondingly, let $\bm{T}_i^{v2t}$ and $\bm{T}_i^{t2v}$ be the row-wise and column-wise normalized refined alignment for the $i$-th sample, respectively. The OT-guided re-aligning is defined as:
\begin{equation}\label{eq: ot_assignment}
\small
\mathcal{L}^{otr} = \frac{1}{N} \sum_{i=1}^{N} \text{KL}(\bm{T}_i^{v2t} \| \bm{P}_i^{v2t}) +  \frac{1}{N+1} \sum_{i=1}^{N+1}\text{KL}(\bm{T}_i^{t2v} \| \bm{P}_i^{t2v}).
\end{equation}
Moreover, we empirically observe that preserving the textual semantic separation term can make the unlearning more stable. Thus, the final unlearning objective is defined as:
\begin{equation}\label{eq: UL_loss}
\mathcal{L}^{UL} = \mathcal{L}^{otr} + \mathcal{L}^{sep} .
\end{equation}

\input{main_tables/zeroshot_classification}

\input{main_tables/zeroshot_retrieval}
\section{Experiment}
\label{sec:experiment}
In this section, we experimentally analyze the effectiveness of NCU in unlearning the NC knowledge from CLIP.

\subsection{Setup}
\paragraph{Datasets.} Our experiments are conducted on three vision-language datasets at different scales and noise: Conceptual Captions 3M (CC3M) \cite{sharma2018conceptual}, Conceptual Captions 12M (CC12M) \cite{changpinyo2021conceptual}, and YFCC15M-R (provided by \cite{gu2024rwkv}, an LLM-recaptioned subset from the YFCC100M \cite{thomee2016yfcc100m}). All datasets are web-crawled and contain an unknown portion of NC pairs, \eg, CC3M is estimated to include at least 3\% false positives. We evaluate NCU on ImageNet and 15 common downstream datasets for classification performance and on MSCOCO and Flickr30K for retrieval capability. Details for datasets are shown in Supplementary \blue{C}.

\paragraph{Unlearning Details.}
Following CLIP, we consider two architectures for the image encoder, \ie, ViT/B16 and ViT/B32, while the text encoder adopts the transformer architecture. We consider a CLIP pre-trained on dataset $\mathbbm{D}$, \eg, CC3M, CC12M, or YFCC15M-R, as our reference model, then we perform the NC unlearning on $\mathbbm{D}$ or its subset to enhance CLIP's robustness. For all experiments, we allocate 2 epochs for learning negative semantics and 8 epochs for noise unlearning. All models are trained with a batch size of 2,048 on 16 NVIDIA V100 GPUs. Detailed training settings are presented in Supplementary \blue{D}.

\paragraph{Evaluation Protocol.} 
We evaluate NCU's transferability with Zero-Shot (ZS) classification accuracy and Linear Probing (LP) accuracy. For ZS classification, we follow CLIP's \cite{radford2021learning} prompt templates to compute distances between class text embeddings and image features. For LP, we follow the mainstream setting \cite{radford2021learning, fan2024improving} that trains a linear classifier using L-BFGS on features extracted from the frozen image encoder. Besides, we evaluate the retrieval performance with the Recall at rank K (R@K) metric.

\subsection{Evaluation on Diverse Downstream Tasks}
To verify the generalization of NCU, we compare it with CLIP on three different types of downstream tasks.

\paragraph{Zero-Shot Transfer.}
We compare the zero-shot performance of CLIP and NCU on $16$ popular image classification
datasets. We follow the prompt templates suggested in the CLIP paper \cite{radford2021learning} to form each class name into a natural sentence. As demonstrated in \cref{tab: zs_classification}, our NCU approach significantly outperforms the baseline CLIP model on both ImageNet and other downstream datasets. Specifically, across all fine-tuning datasets and all model architectures, NCU gains in the range of $2.8 \% \sim 4.1\%$ in top-1 accuracy on ImageNet and $2.5 \% \sim 4.0\%$ on average over the other downstream datasets. This reveals that NCU can successfully eliminate the impact of NC on CLIP by robust fine-tuning with the same dataset.

\paragraph{Image-Text Retrieval.}
We present the zero-shot cross-modal retrieval performance on the testing set of Flickr30K (1K) and MSCOCO (5K) in \cref{tab: zs_retrieval}. Our method considerably outperforms the vanilla CLIP in almost all cases. For instance, when fine-tuning CLIP (ViT-B/32) pre-trained on the CC3M dataset, our NCU method achieves a $7.7\%$ improvement in average recall scores on Flickr30K and $4.8\%$ improvement in average recall scores on MSCOCO. This finding indicates that NCU can remarkably enhance the alignment of images and text in the embedding space.

\paragraph{Linear Probing.}
\cref{tab: linear_probing} reports the linear probing performance on $4$ representative downstream datasets. Our NCU consistently surpasses CLIP in the vast majority of cases, suggesting that the visual embeddings learned by our NCU are more effective and transferable than CLIP.

\input{main_tables/linear_probing}

\subsection{Compared to Robust Methods}
In this section, we compare NCU with other robust-designed techniques against NC on zero-shot ImageNet1K classification task, \ie, gradient ascent (GA), and SoftCLIP \cite{gao2024softclip}. Specifically, we evaluate GA as a standalone method, where $-\mathcal{L}_{CLIP}$ is performed on $\mathbbm{D}_{FG}$ for handling FPs and $\mathcal{L}_{CLIP}$ with label smoothing is performed on $\mathbbm{D}_{RT}$ for FNs. SoftCLIP is a noise-robust SOTA method that trains CLIP from scratch by additional intra-modal guided alignment, \ie, ROI features. As shown in \cref{tab: robust_methods}, although GA is a naive unlearning strategy, it still achieves observable performance gains. Meanwhile, SoftCLIP's self-similarity modeling fails to excavate supervision from false positives, which may explain why it performs worse than GA in some cases, \eg, CC12M with ViT-B/32. By contrast, our NCU achieves solid improvements by forgetting both false positives and false negatives, outperforming SoftCLIP by $1.1\% \sim 2.3\%$ without external guidance.

\input{main_tables/robust_method}
\setlength{\textfloatsep}{16pt}

\subsection{Ablation Study}
To investigate the effectiveness of specific components in our method, we carry out some ablation studies on ImageNet1K with models unlearned on CC3M. We first ablate the contributions of two key components of NCU, \ie, negative prompt and text relation opposite. Specifically, in the variant $\mathcal{V}_1$, we replace the learnable prompt tokens by prepending some textual negative prefixes to raw captions, \eg, \textit{`the image has no'} or \textit{`this picture lacks'}. In the variant $\mathcal{V}_2$, we use maximal $L_2$ distance loss \cite{wang2023clipn} as a substitute for our text relation opposite. Besides, we validate the impact of different NC unlearning by intervening with the refined alignment, \ie, $\mathcal{V}_3$ and $\mathcal{V}_4$. As shown in \cref{tab: ablation}, we observe that: 1) Using negative textual prefixes also shows competitive results, demonstrating the generalization of our method. However, we argue that the learnable prompt is preferable, except for performance gains, operating to features makes it possible for NCU to extend to modalities beyond text. 2) Simply maximizing the distance between negative and original text embeddings leads to suboptimal performance, which aligns with our analysis in \cref{fig: illustration}. 3)  Both types of NC impair CLIP's performance, among which the false positive causes a more severe impact. While NCU achieves the best performance by forgetting such noisy knowledge.
\input{main_tables/ablation}

\vspace{0.2cm}
\begin{figure}[h]
\centering  
\includegraphics[width=\columnwidth]{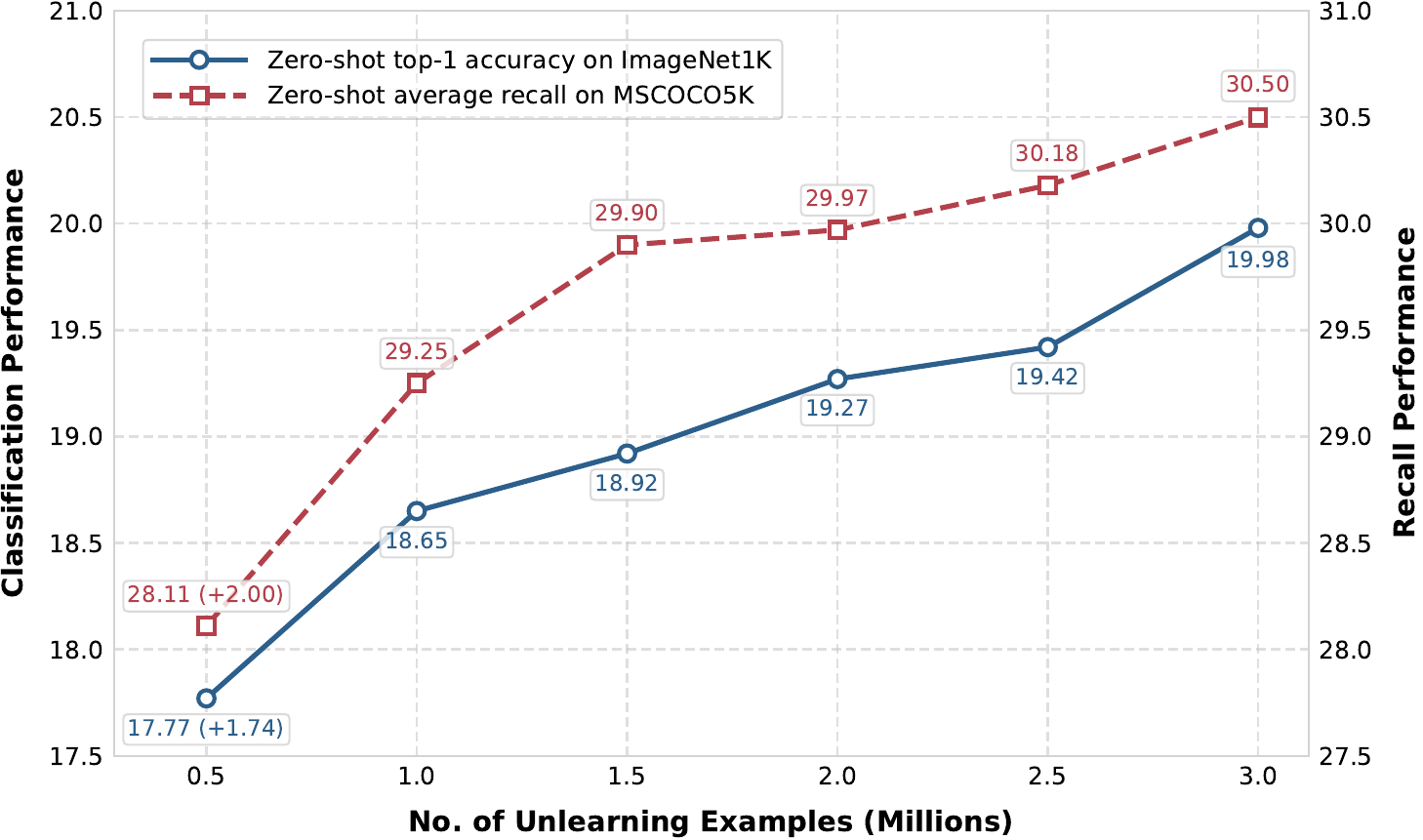}
\caption{Effect of NCU with varying fine-tuning dataset sizes on zero-shot image classification and cross-modal retrieval.}
% \vspace{-0.1cm}
\label{fig: partial_UL}
\end{figure}

\subsection{NC Unlearning with Partial Data} \label{subsec: un_partial}
In this section, we conduct an interesting study to verify whether CLIP can improve robustness by only unlearning NC with a portion of the pre-trained data. To this end, given a CLIP pre-trained on CC3M as the reference model, we evaluate NCU using different data portions ranging from 0.5 million to 3 million image-text pairs. \cref{fig: partial_UL} plots the ZS top-1 accuracy on ImageNet1K and the average of recalls on MSCOCO 5K. Remarkably, even when unlearning on less than 20\% of the original data (0.5M), NCU achieves significant performance gains while preserving overall knowledge learned in CC3M. With the accessible data increasing, NCU shows consistent improvements on both zero-shot downstream tasks. This phenomenon indicates NCU's flexibility in enhancing the robustness of models with limited data, which is valuable to handling VLMs pre-trained with partially private or proprietary data.

% which is particularly valuable for VLMs pre-trained with private or proprietary data.
% This phenomenon indicates NCU's flexibility in enhancing model robustness with limited data, which is particularly.

% \vspace{10pt}
\subsection{Visualization and Analysis}
\begin{figure}[t]
\centering
\subfloat[NCU]{\includegraphics[width=0.5\columnwidth]{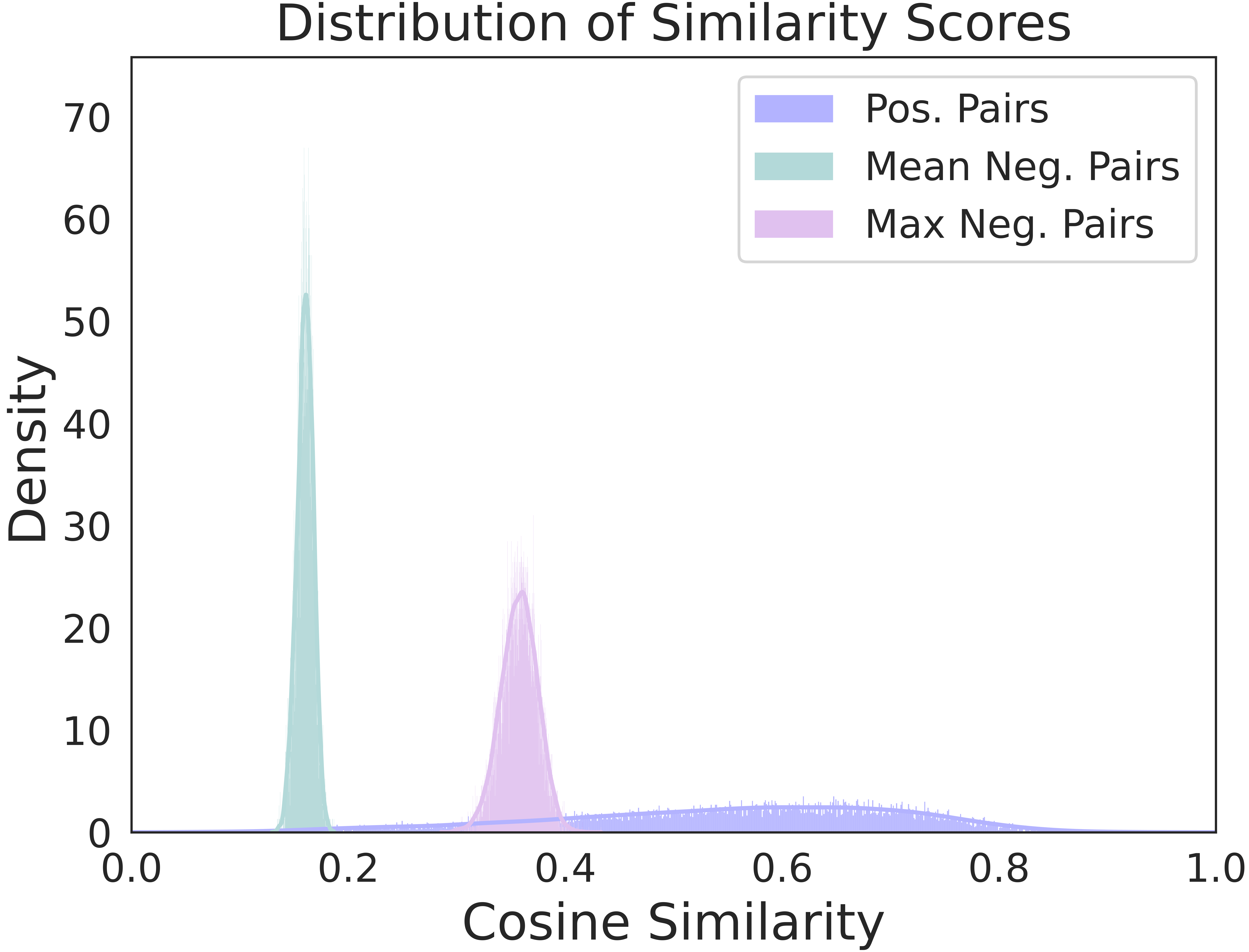}%
\label{fig: NCU_sims}}
\subfloat[CLIP]{\includegraphics[width=0.5\columnwidth]{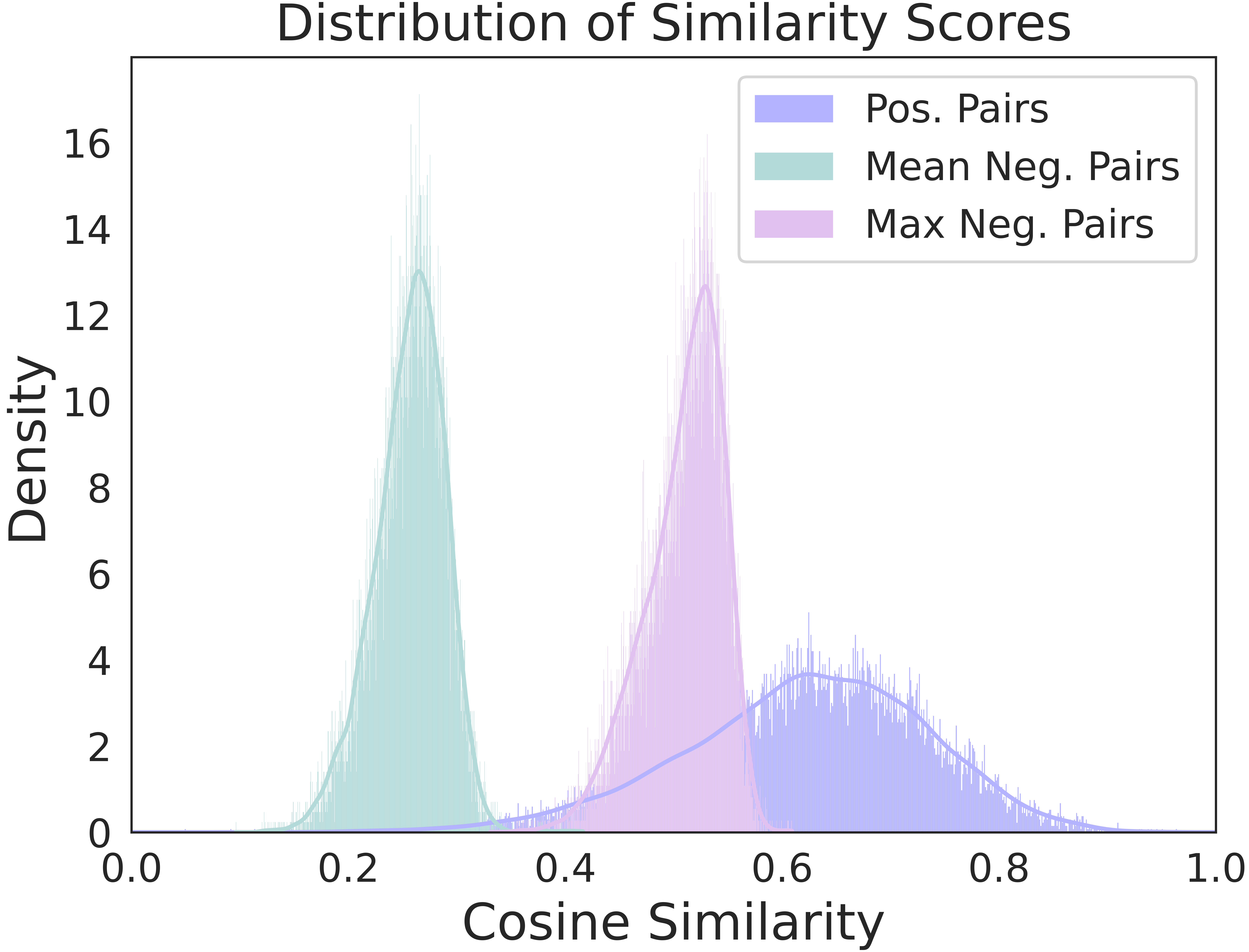}%
\label{fig: CLIP_sims}}
\caption{Similarity scores distribution of positive and negative pairs from CLIP and NCU. Both models are based on ViT/B16 and learning from the CC3M training set.}
\label{fig: sims_dist}
\end{figure}

To intuitively show the robust embedding space that is refined by our approach, we plot the distribution of normalized similarity for CLIP and NCU on the validation set of CC3M. In \cref{fig: sims_dist}, we illustrate similarity scores for positive pairs, mean of negative pairs, and top $5\%$ maximum of negative pairs. First, we observe that NCU produces a wider distribution of positive similarity scores, capturing more fine-grained matching degrees among positive pairs. Second, NCU improves the feature discrimination, which leads to a more significant separation between positive and negative pairs. 
Lastly, NCU provides more appropriate measures for hard negatives, which maintains separation from both positive and other negative pairs.

\section{Limitations and Future Works}
Our work still has certain limitations due to the finite computing capability, including 1) This work only uses CLIP to explore the efficacy of NCU. Further research is needed to confirm its applicability in other VLMs, such as BLIP-2 \cite{li2023blip}, and even larger VLMs like VisionLLM \cite{wang2023visionllm} or InternVL \cite{chen2024internvl}. 2) The current experiments are mainly conducted on million-scale data, and we plan to extend it to larger-scale datasets to verify NCU's generalization.

\section{Conclusion}
This work provides a new thinking in robust vision-language learning. Instead of re-training models from scratch, we suggest eliminating the harmful effects of noisy correspondence from pre-trained models. To this end, we propose NCU, a robust fine-tuning framework that efficiently unlearns noisy correspondence in CLIP. Our key concept is to learn the hardest negative information that can provide explicit unlearning direction to resist both FP and FN. We formalize such twin goals unlearning process into one unified OT problem for fast fine-tuning. Extensive experiments are conducted to verify that NCU can endow CLIP with strong robustness against noisy correspondence.

\section*{Acknowledgments}
This work was supported in part by the Major Key Project of PCL under Grant PCL2025AS10 and PCL2024A06, and in part by the Shenzhen Science and Technology Program under Grant RCJC20231211085918010.

% Project of China Knowledge Center for Engineering Science and Technology, and Project of Chinese academy of engineering “The Online and Offline Mixed Educational Service System for ‘The Belt and Road’ Training in MOOC China”. We would like to express our gratitude for the support of K.C. Wong Education Foundation.
% \end{comment}

%%%%%%%%% REFERENCES
{\small
\bibliographystyle{ieee_fullname}
\bibliography{egbib}
}
% \appendix
% \appendixtitle
% \input{Appendix}
\end{document}

%% file: main_tables/zeroshot_classification.tex
\begin{table*}[t]
  \centering
  % \addtolength{\tabcolsep}{-1pt}
  \renewcommand\arraystretch{1}
  \resizebox{1.0\textwidth}{!}{
    \begin{tabular}{cc|cccccccccccccccl|l}
   \hline
   \hline
   \raisebox{4ex}{ Dataset}  & \raisebox{4ex}{Model}  & \rotatebox[origin=lb]{90}{\smash{\small Caltech101}} & \rotatebox[origin=lb]{90}{\smash{\small CIFAR-10}} & \rotatebox[origin=lb]{90}{\smash{\small CIFAR-100}} & \rotatebox[origin=lb]{90}{\smash{\small DTD}} & \rotatebox[origin=lb]{90}{\smash{\small Aircraft}} & \rotatebox[origin=lb]{90}{\smash{\small SST2}} & \rotatebox[origin=lb]{90}{\smash{\small Flowers102}} & \rotatebox[origin=lb]{90}{\smash{\small Food101}} & \rotatebox[origin=lb]{90}{\smash{\small GTSRB}} & \rotatebox[origin=lb]{90}{\smash{\small OxfordPets}} & \rotatebox[origin=lb]{90}{\smash{\small RESISC45}} & \rotatebox[origin=lb]{90}{\smash{\small SUN397}} & \rotatebox[origin=lb]{90}{\smash{\small EuroSAT}} & \rotatebox[origin=lb]{90}{\smash{\small StanfordCars}} & \rotatebox[origin=lb]{90}{\smash{\small STL10}} & \multicolumn{1}{l|}{\quad \ \rotatebox[origin=lb]{90}{\smash{\small \bf Average}}} & \multicolumn{1}{l}{\quad \ \rotatebox[origin=lb]{90}{\smash{\small \bf ImageNet1K}}} \\
    \hline
    \multicolumn{19}{c}{\textit{Model Architecture: ViT-B/16}} \\
    \hline
    \multirow{2.2}[0]{*}{CC3M} & CLIP  & 52.3  & \textbf{55.2} & 24.1  & 10.9  & 1.0   & \textbf{50.1} & 11.9  & 11.1  & 6.9   & 12.9  & 19.5  & 25.0  & 13.5  & 0.8   & 81.7  & 25.1  & 16.0 \\

          & NCU & \textbf{59.1} & 54.3  & \textbf{28.8} & \textbf{12.3} & \textbf{1.1} & \textbf{50.1} & \textbf{14.1} & \textbf{14.8} & \textbf{7.4} & \textbf{16.3} & \textbf{22.8} & \textbf{32.3} & \textbf{21.7} & \textbf{1.5} & \textbf{86.3} &  \textbf{28.2}\green{$\uparrow_{3.1}$} & \textbf{20.0}\green{$\uparrow_{4.0}$} \\
          \hline
    \multirow{2}[1]{*}{CC12M} & CLIP  &    77.0  & 66.5  & 38.3  & 21.2  & 2.5   & 47.7  & \textbf{33.4} & 51.9  & 7.3   & 64.2  & 39.0  & 44.7  & 21.2  & 25.5  & 91.4  & 42.1  & 40.6 \\
          & NCU & 
    \textbf{80.9} & \textbf{79.3} & \textbf{49.1} & \textbf{23.2} & \textbf{2.7} & \textbf{48.0} & 31.7  & \textbf{52.7} & \textbf{10.1} & \textbf{66.5} & \textbf{41.9} & \textbf{52.6} & \textbf{28.6} & \textbf{29.0} & \textbf{93.2} &  \textbf{46.0}\green{$\uparrow_{3.9}$} & \textbf{43.4}\green{$\uparrow_{2.8}$}\\
    \hline
    \multicolumn{19}{c}{\textit{Model Architecture: ViT-B/32}} \\
    \hline
    \multirow{1.8}[2]{*}{CC3M} & CLIP  & {47.7} & {54.2} & {18.0} & {7.6} & {1.2} & {\textbf{50.1}} & {9.3} & {9.1} & {6.0} & {7.4} & {16.2} & {16.0} & {15.5} & {0.8} & {77.7} & {22.5} & {11.8} \\
          & NCU & {\textbf{53.0}} & {\textbf{56.7}} & {\textbf{25.9}} & {\textbf{10.4}} & {\textbf{1.7}} & {\textbf{50.1}} & {\textbf{10.2}} & {\textbf{10.5}} & {\textbf{6.5}} & {\textbf{10.5}} & {\textbf{19.0}} & {\textbf{22.2}} & {\textbf{16.7}} & {\textbf{1.4}} & {\textbf{80.1}} &  {\textbf{25.0}}\green{$\uparrow_{2.5}$} & {\textbf{14.6}}\green{$\uparrow_{2.8}$} \\
    \hline
    \multirow{1.8}[2]{*}{CC12M} & CLIP  & 76.3  & 68.2  & 35.2  & 16.1  & \textbf{2.8} & 50.1  & \textbf{29.3} & 37.6  & 6.4   & 54.1  & 30.1  & 39.2  & 22.5  & 14.8  & 90.8  & 38.2  & 33.8 \\
          & NCU & 
    \textbf{80.4} & \textbf{68.5} & \textbf{41.4} & \textbf{19.3} & 2.6   & \textbf{52.8} & 28.6  & \textbf{43.4} & \textbf{7.2} & \textbf{62.4} & \textbf{35.7} & \textbf{48.3} & \textbf{31.3} & \textbf{18.1} & \textbf{92.5} &  \textbf{42.2}\green{$\uparrow_{4.0}$} & \textbf{36.7}\green{$\uparrow_{2.9}$} \\
    \hline
    \multirow{1.7}[2]{*}{YFCC15M-R} & CLIP  & 53.7  & 67.0  & 34.4  & 13.1  & 1.1   & 49.3  & 22.1  & 18.6  & 11.0   & 13.5  & 20.3  & 29.3  & 23.0 & 1.7   & 83.7  & 29.5  & 17.8 \\

          & NCU &    \textbf{58.2} & \textbf{69.5} & \textbf{37.8} & \textbf{15.3} & \textbf{1.8} & \textbf{49.9} & \textbf{29.2} & \textbf{23.7} & \textbf{11.2} & \textbf{16.1} & \textbf{23.1} & \textbf{34.0} & \textbf{23.3}  & \textbf{1.8} & \textbf{86.7} &  \textbf{32.1}\green{$\uparrow_{2.6}$} & \textbf{21.9}\green{$\uparrow_{4.1}$} \\

    \hline
    \hline
    \end{tabular}}

  \caption{Zero-shot transfer evaluation of different models.}
    \label{tab: zs_classification}%
    \vspace{6pt}
\end{table*}%

%% file: main_tables/zeroshot_retrieval.tex
\begin{table*}[t]
  \centering
  
  \renewcommand\arraystretch{1}
  \addtolength{\tabcolsep}{-2pt}
  \resizebox{1.0\textwidth}{!}{
    \begin{tabular}{c|c|c|cccccc|l|cccccc|l}
    \hline
    \hline
    \multicolumn{1}{c|}{\multirow{2.2}[4]{*}{\newline{}Dataset\newline{}}} & \multirow{2.2}[4]{*}{Architecture} & \multirow{2.2}[4]{*}{Model } & \multicolumn{7}{c|}{Flickr30K 1K Testing}             & \multicolumn{7}{c}{MSCOCO 5K Testing} \\
\cline{4-17}          &       &       & \multicolumn{3}{c}{Image-to-Text} & \multicolumn{3}{c|}{Text-to-Image} & \multicolumn{1}{c|}{\multirow{1.8}[2]{*}{\textbf{Average}}} & \multicolumn{3}{c}{Image-to-Text} & \multicolumn{3}{c|}{Text-to-Image} & \multicolumn{1}{c}{\multirow{1.8}[2]{*}{\textbf{Average}}} \\
          &       &       & R@1   & R@5   & R@10  & R@1   & R@5   & R@10  &       & R@1   & R@5   & R@10  & R@1   & R@5   & R@10  &  \\
    \hline
    \multirow{3}[4]{*}{CC3M} & \multirow{1.8}[2]{*}{ViT-B/16} & CLIP  & 27.6  & 54.2  & 65.8  & 19.0  & 40.5  & 51.3  & \ \ \ 43.1  & 12.8  & 30.9  & 42.7  & 9.7   & 25.4  & 35.2  & \ \ \ 26.1 \\
          &       & NCU   & \textbf{32.1} & \textbf{60.7} & \textbf{71.9} & \textbf{25.2} & \textbf{49.5} & \textbf{61.0} & \textbf{\ \ \ 50.1}{\green{${\uparrow_{7.0}}$}} & \textbf{15.8} & \textbf{36.9} & \textbf{48.4} & \textbf{12.1} & \textbf{29.7} & \textbf{40.1} & \textbf{\ \ \ 30.5}\green{$\uparrow_{{4.4}}$} \\
\cline{2-17}          & \multirow{2}[2]{*}{ViT-B/32} & CLIP  & 14.0  & 35.2  & 47.7  & 11.5  & 27.9  & 37.8  & \ \ \ 29.0  & 7.0   & 20.1  & 28.9  & 6.0   & 16.7  & 24.0  & \ \ \ 17.1 \\
          &       & NCU   & \textbf{21.3} & \textbf{44.5} & \textbf{55.7} & \textbf{15.7} & \textbf{36.3} & \textbf{46.4} & \textbf{\ \ \ 36.7}\green{$\uparrow_{{{7.7}}}$}& \textbf{10.7} & \textbf{25.7} & \textbf{35.4} & \textbf{8.1} & \textbf{21.3} & \textbf{30.2} & \textbf{\ \ \ 21.9}\green{$\uparrow_{{4.8}}$} \\
    \hline
    \multirow{2}[2]{*}{CC12M} & \multirow{2}[2]{*}{ViT-B/32} & CLIP  & 50.3  & 77.2  & \textbf{85.9} & 37.9  & 64.8  & 74.2  & \ \ \ 65.1  & 26.8  & 54.0  & 65.9  & \textbf{20.0} & 42.4  & 54.2  & \ \ \ 43.9 \\
          &       & NCU   & \textbf{53.0} & \textbf{77.3} & 85.0  & \textbf{38.4} & \textbf{66.6} & \textbf{76.7} & \textbf{\ \ \ 66.2}\green{$\uparrow_{{1.1}}$} & \textbf{28.2} & \textbf{54.7} & \textbf{66.7} & \textbf{20.0} & \textbf{42.8} & \textbf{54.3} & \textbf{\ \ \ 44.5}\green{$\uparrow_{{0.6}}$}\\
    \hline
    \multirow{2}[2]{*}{YFCC15M-R} & \multirow{2}[2]{*}{ViT-B/32} & CLIP  & 57.4  & 81.6  & 89.1  & 40.8  & 66.4  & 75.4  & \ \ \ 68.5  & \textbf{35.0}  & 60.9  & 71.8  & 22.6  & 46.0  & 58.0  & \ \ \ 49.1 \\
          &       & NCU   & \textbf{58.0} & \textbf{83.2} & \textbf{89.7} & \textbf{42.5} & \textbf{69.9} & \textbf{78.8} & \textbf{\ \ \ 70.4}\green{$\uparrow_{{1.9}}$} & 34.3 & \textbf{62.0} & \textbf{73.5} & \textbf{24.6} & \textbf{49.0} & \textbf{60.8} & \textbf{\ \ \ 50.7}\green{$\uparrow_{{1.6}}$} \\
    \hline
    \hline
    \end{tabular}%
    }

  \caption{Zero-shot cross-modal retrieval evaluation of different models.}
      \label{tab: zs_retrieval}%
\end{table*}%

%% file: main_tables/linear_probing.tex
\begin{table}[t]
  \centering
    % \renewcommand\arraystretch{0.99}
    % \small
    \addtolength{\tabcolsep}{-1.5pt}  
    \renewcommand{\arraystretch}{1}
    \resizebox{1.0\columnwidth}{!}{
    \begin{tabular}{c|c|c|cccc}
    \hline
    \hline
   \multirow{-3}[1]{*}{Dataset} & \multirow{-3}[1]{*}{Architecture} & \multirow{-3}[1]{*}{Model} & \rotatebox[origin=lb]{60}{\small{SUN397}}  & \rotatebox[origin=lb]{60}{\small{OxfordPets}} & \rotatebox[origin=lb]{60}{\small{Food101}} & \rotatebox[origin=lb]{60}{\small{ImageNet}} \\
    \hline
    \multirow{3.5}[4]{*}{CC3M} & \multirow{2}[2]{*}{ViT-B/16} & CLIP  & 54.38 & 62.20 & 54.36 & 48.13 \\
          &       & NCU   & \textbf{55.60} & \textbf{62.85} & \textbf{54.74} & \textbf{49.90} \\
\cline{2-7}          & \multirow{2}[2]{*}{ViT-B/32} & CLIP  & 46.96 & 52.30 & \textbf{46.96} & 40.20 \\
          &       & NCU   & \textbf{48.20} & \textbf{52.82} & 46.85 & \textbf{41.49} \\
    \hline
    \multirow{3.5}[4]{*}{CC12M} & \multirow{2}[2]{*}{ViT-B/16} & CLIP  & 70.94 & 82.83 & 78.99 & \textbf{66.70} \\
          &       & NCU   & \textbf{71.36} & \textbf{84.76} & \textbf{79.18} & 66.65 \\
\cline{2-7}          & \multirow{2}[2]{*}{ViT-B/32} & CLIP  & 66.05 & 78.41 & 70.12 & 59.19 \\
          &       & NCU   & \textbf{66.63} & \textbf{78.63} & \textbf{70.76} & \textbf{60.34} \\
    \hline
    \multirow{1.8}[2]{*}{YFCC15M-R} & \multirow{2}[2]{*}{ViT-B/32} & CLIP  & 60.46 & 61.90 & 59.09 & 51.07 \\
          &       & NCU   & \textbf{60.75} & \textbf{62.85} & \textbf{60.46} & \textbf{52.29} \\
    \hline
    \hline
    \end{tabular}%
    }

  \caption{Linear probing comparison of different models.}
  \vspace{0.1cm}
  \label{tab: linear_probing}%
\end{table}%

%% file: main_tables/robust_method.tex
\begin{table}[h]
  \centering
    % \renewcommand\arraystretch{0.95}
    % \small
    \addtolength{\tabcolsep}{3.6pt}  
    \renewcommand{\arraystretch}{1}
    \resizebox{1.0\columnwidth}{!}{
    \begin{tabular}{c|l|c|l}
    \hline
    \hline
    \multirow{1.8}[2]{*}{Dataset} & \multirow{1.8}[2]{*}{Model} & Model  & \ ImageNet1K \\
          &       & Architecture &  \ \ \ \ ZS top-1 \\
    \hline
    \multirow{3.8}[2]{*}{CC3M} & CLIP  & \multirow{3.8}[2]{*}{ViT-B/16} & \ \ \ \ \ 16.0 \\
          & Gradient Ascent &       & \ \ \ \ \ 16.7 \green{$\uparrow_{0.7}$} \\
          & SoftCLIP &       & \ \ \ \ \ 18.9 \green{$\uparrow_{2.9}$} \\
          & NCU   &       & \ \ \ \ \ \textbf{20.0} \green{$\uparrow_{4.0}$} \\ 
    
    \hline
    \multirow{3.8}[2]{*}{CC3M} & CLIP  & \multirow{3.8}[2]{*}{ViT-B/32} & \ \ \ \ \ 11.8 \\
          & Gradient Ascent &       & \ \ \ \ \ 12.1 \green{$\uparrow_{0.3}$}\\
          & SoftCLIP &       & \ \ \ \ \ 13.3 \green{$\uparrow_{1.5}$}\\
          & NCU   &       & \ \ \ \ \ \textbf{14.6} \green{$\uparrow_{2.8}$} \\
    \hline
    \multirow{3.8}[2]{*}{CC12M} & CLIP  & \multirow{3.8}[2]{*}{ViT-B/16} & \ \ \ \ \ 40.6 \\
          & Gradient Ascent &       & \ \ \ \ \ 41.6 \green{$\uparrow_{1.0}$}\\
          & SoftCLIP &       & \ \ \ \ \ 42.1 \green{$\uparrow_{1.5}$}\\
          & NCU   &       & \ \ \ \ \ \textbf{43.4} \green{$\uparrow_{2.8}$}\\
    \hline
    \multirow{3.8}[2]{*}{CC12M} & CLIP  & \multirow{3.8}[2]{*}{ViT-B/32} & \ \ \ \ \ 33.8 \\
          & Gradient Ascent &       & \ \ \ \ \ 35.1 \green{$\uparrow_{1.3}$}\\
          & SoftCLIP &       & \ \ \ \ \ 34.4 \green{$\uparrow_{0.6}$}\\
          & NCU   &       & \ \ \ \ \ \textbf{36.7} \green{$\uparrow_{2.9}$}\\
    \hline
    \hline
    \end{tabular}%
    }

  \caption{Zero-shot top-1 performance on ImageNet1K.}
  \label{tab: robust_methods}%
\end{table}%

%% file: main_tables/ablation.tex
\begin{table}[t]
  \centering
  \resizebox{1\columnwidth}{!}{
    %  \renewcommand\arraystretch{.9}
    % \small
    \addtolength{\tabcolsep}{-0pt}  
    \renewcommand{\arraystretch}{1}
    \begin{tabular}{l|ll}
    \hline
    \hline
    \multirow{2}[1]{*}{Model} & \multicolumn{2}{c}{ImageNet1K ZS top-1} \\
    \cline{2-3}          & \ \ ViT-B/16 & \ \ ViT-B/32 \\
    \hline
    NCU   & \ \ \ \ \textbf{20.0}  & \ \ \ \ \textbf{14.6} \\
    $\mathcal{V}_1$ (w/o Hardest Negative Prompts) & \ \ \ \ 19.3 \green{$\downarrow_{0.7}$}   & \ \ \ \ 14.4 \green{$\downarrow_{0.2}$}  \\
    $\mathcal{V}_2$ (w/o Text Relation Opposite) & \ \ \ \ 18.8 \green{$\downarrow_{1.2}$}  & \ \ \ \ 13.9 \green{$\downarrow_{0.7}$} \\
    $\mathcal{V}_3$ (w only False Negatives Unlearning) & \ \ \ \ 17.7 \green{$\downarrow_{2.3}$}  & \ \ \ \ 13.5 \green{$\downarrow_{1.1}$}\\
    $\mathcal{V}_4$ (w only False Positives Unlearning) & \ \ \ \ 19.2 \green{$\downarrow_{0.8}$} & \ \ \ \ 14.3 \green{$\downarrow_{0.3}$} \\
    \hline
    \hline
    \end{tabular}%
    }
  \caption{Ablation studies on zero-shot transfer task (ImageNet1K) of models unlearned on CC3M.}
  \label{tab: ablation}%
\end{table}%